\documentclass[10pt,twocolumn,letterpaper]{article}

\usepackage{pdfpages}

\usepackage{cvpr}              
\definecolor{cvprblue}{rgb}{0.21,0.49,0.74}
\usepackage[pagebackref,breaklinks,colorlinks,allcolors=cvprblue]{hyperref}

\usepackage{bm}
\usepackage{float}
\usepackage{multirow}

\def\paperID{10207}
\def\confName{CVPR}
\def\confYear{2026}

\title{Towards Realistic and Consistent Orbital Video Generation via \\ 3D Foundation Priors}

\author{
Rong Wang$^{1 *}$ \quad
Ruyi Zha$^{1}$ \quad
Ziang Cheng$^{2}$ \quad
Jiayu Yang$^{2}$ \quad
Pulak Purkait$^{2}$ \quad
Hongdong Li$^{1,2 *}$ \\
\vspace{0.3cm}
$^1$Australian National University \hspace{0.4cm} $^2$Amazon
}

\begin{document}

\twocolumn[{%
\renewcommand\twocolumn[1][]{#1}%
\maketitle
\begin{center}
    \centering
    \vspace{-0.6cm}
    \includegraphics[width=\textwidth]{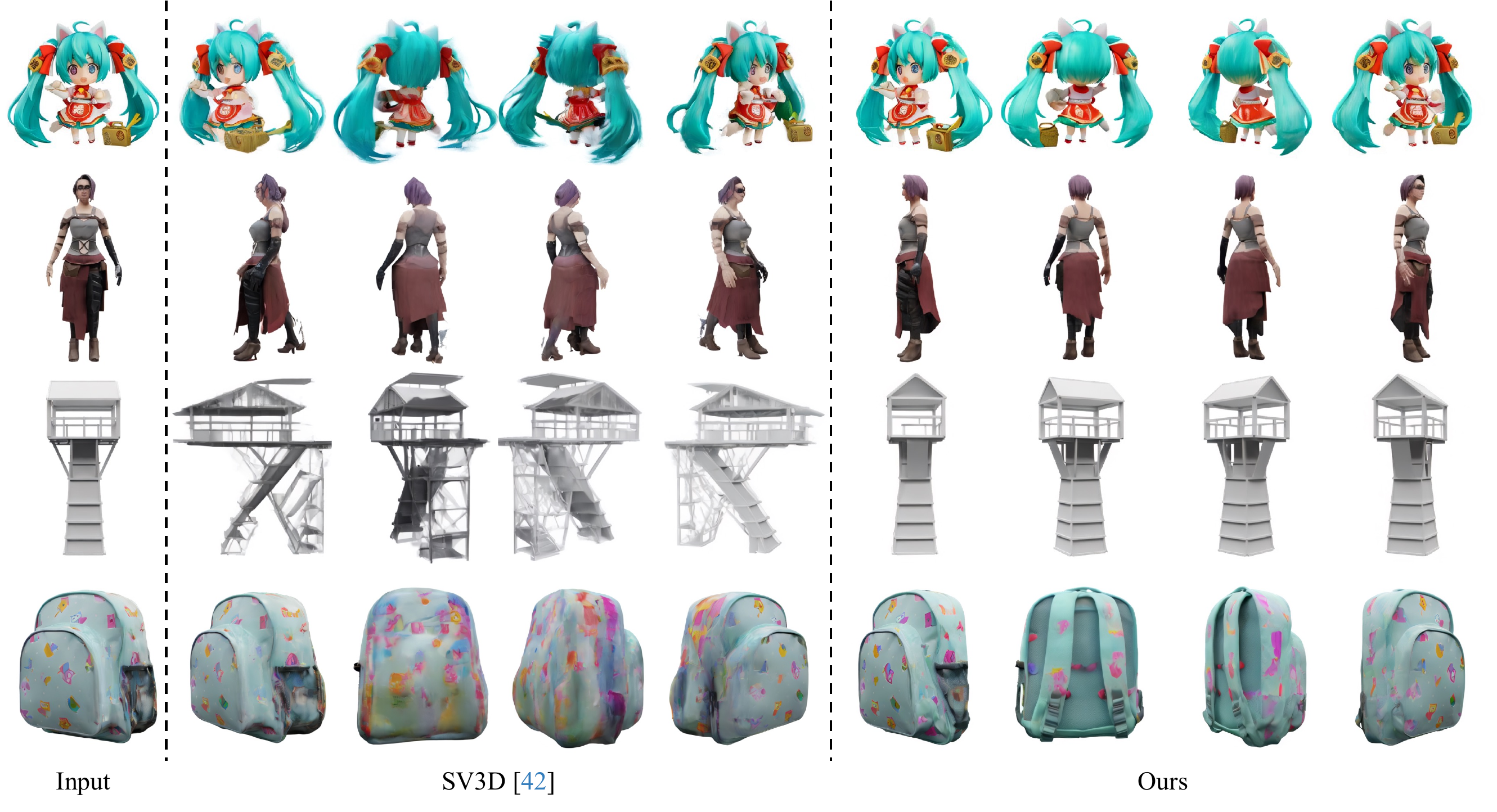}
    \vspace{-0.6cm}\captionsetup{type=figure}
    \captionof{figure}{In this work, we generate realistic and consistent orbital videos from a single object image. 
    By leverage shape priors from 3D foundation model, our method can robustly produce consistent frames with more \emph{plausible} structures compared to the baseline \cite{voleti2024sv3d}.}
    \label{open}
\end{center}%
}]
\maketitle

\renewcommand{\thefootnote}{*}
\footnotetext{Work done while Rong Wang was a research intern at Amazon. HL is funded in part by ARC Grant DP220100800.}

\begin{abstract}
We present a novel method for generating geometrically realistic and consistent orbital videos from a single image of an object. Existing video generation works mostly rely on pixel-wise attention to enforce view consistency across frames. However, such mechanism does not impose sufficient constraints for long-range extrapolation, \emph{e.g.} rear-view synthesis, in which pixel correspondences to the input image are limited. Consequently, these works often fail to produce results with a plausible and coherent structure. To tackle this issue, we propose to leverage rich shape priors from a 3D foundational generative model as an auxiliary constraint, motivated by its capability of modeling realistic object shape distributions learned from large 3D asset corpora. Specifically, we prompt the video generation with two scales of latent features encoded by the 3D foundation model: (i) a denoised global latent vector as an overall structural guidance, and (ii) a set of latent images projected from volumetric features to provide view-dependent and fine-grained geometry details. In contrast to commonly used 2.5D representations such as depth or normal maps, these compact features can model complete object shapes, and help to improve inference efficiency by avoiding explicit mesh extraction. To achieve effective shape conditioning, we introduce a multi-scale 3D adapter to inject feature tokens to the base video model via cross-attention, which retains its capabilities from general video pretraining and enables a simple and model-agonistic fine-tuning process. Extensive experiments on multiple benchmarks show that our method achieves superior visual quality, shape realism and multi-view consistency compared to state-of-the-art methods, and robustly generalizes to complex camera trajectories and in-the-wild images. 
\end{abstract}

\section{Introduction}

\noindent
Generating realistic videos from an object image and camera trajectory enables a wide range of applications, including e-commerce, extended reality, and game design. To this end, it has received increasing attention \citep{kwak2024vivid, chen2024v3d, voleti2024sv3d, yang2024hi3d}. A key challenge in this task is to ensure all generated frames agree on a coherent and plausible object shape, especially for unseen objects and challenging viewing angles. To this end, we aim to generate consistent video frames with realistic object shapes in this work, thereby improving the model robustness desired by downstream applications.

Existing video generation works \cite{blattmann2023stable, blattmann2023align, luo2023videofusion, yang2024cogvideox} have demonstrated impressive quality and fidelity by scaling large diffusion transformers \cite{peebles2023scalable}. While self-attention layers \cite{vaswani2017attention} in these transformer blocks can model view consistency between actively matching patches, they are ineffective for learning dependencies between frames with large viewpoint changes, \emph{e.g.} input front view and target rear view, where pixel-wise correspondences are limited and unreliable. Without further 3D world knowledge, these methods often produce unrealistic results, containing artifacts of distorted and unnatural shapes as illustrated in Figure \ref{open}. Meanwhile, recent works \cite{ren2025gen3c, muller2024multidiff, cao2024mvgenmaster} attempt to incorporate geometry conditions from 2D foundation models \cite{bhat2023zoedepth, yang2024depth}, \emph{e.g.} one-view depth maps, to align the video generation. However, such 2.5D priors can not model complete object shapes, hence generation for unobserved or occluded object parts remains under-constrained.

In this work, we present a novel method to effectively improve shape realism and view consistency for orbital video generation via \emph{3D foundation priors}, which is achieved by leveraging the power of recently developed 3D generative models \cite{li2024craftsman, zhang2024clay, zhao2025hunyuan3d}. These 3D foundation models learn to directly reconstruct complete object shapes in a \emph{native} 3D latent space, and thus are capable of encoding rich object structural information and modeling realistic object shape distributions after being trained with large-scaled 3D asset corpora and explicit geometry supervision. A key observation is that their latent features can be used as an effective 3D shape prior, which not only provides auxiliary constraints for synthesizing challenging target views, but also enhances view consistency thanks to a unified 3D reference. To this end, we propose to prompt the video generation with an additional shape condition, \emph{i.e.} two scales of latent features encoded by the 3D foundation model as: (\emph{i}) a global latent vector acts as an overall structural guidance, and (\emph{ii}) a set of latent images projected from volumetric features queried on the global vector, which provides fine-grained and view-dependent geometry details. By using compact latent features as condition signals, we avoid time-consuming mesh extraction process and ensure an efficient inference pipeline. Furthermore, we introduce a multi-scale 3D adapter with alternating cross attention layers to effectively inject both features to the base video generation model. In this way, the adapter can be efficiently and flexibly plugged into any base video transformers, while retaining its capability inherited from general video pretraining. We train our model on a large-scale dataset with synthetic videos rendered from high-quality assets, which ensures model generalizability across diverse object types.

Our contributions can be summarized as follows. (\emph{i}) We propose a novel method that incorporates the video generation pipeline with 3D foundation priors, which noticeably improves shape realism and view consistency in generated results. (\emph{ii}) We introduce a multi-scale 3D adapter, which effectively and efficiently injects latent shape features into the base video models to prompt video generation with 3D shape conditions. Extensive experiments on multiple benchmarks show that our method generates results with more realistic shapes and consistent frames compared to various baseline methods, and consistently applies to complex camera trajectories and in-the-wild examples.

\section{Related Works}

\textbf{Novel View Synthesis. }Recently developed diffusion models \cite{podell2023sdxl, song2020denoising, rombach2022high} have shown promising success in generating high-quality images under various text and image conditioning. In view of this, several works \cite{liu2023zero, shi2023mvdream, kong2024eschernet, zheng2023free3d, liu2023syncdreamer, long2024wonder3d, li2024era3d} have extended them for novel view synthesis (NVS), which is formulated as camera-conditioned image generation. Early works \cite{liu2023zero, objaverseXL} mostly condition on coarse camera prompts such as relative pose differences, and independently generate individual frames, thus often suffering inaccurate camera control and inconsistency across views. In contrast, \cite{zheng2023free3d, kong2024eschernet} then propose to refine the camera condition with more precise ray \cite{sitzmann2021light} or position embeddings for more accurate camera control. Moreover, most works \cite{yang2024consistnet, long2024wonder3d, li2024era3d, shi2023mvdream, liu2023syncdreamer} observe that pixel-wise attention is critical for enforcing view consistency between concurrently generated frames, which can be specialized as epipolar \cite{huang2024epidiff, yang2024consistnet} or scanline \cite{li2024era3d} attention for memory efficiency. However, all above works only make
use of image diffusion models and generate \emph{sparse} frames. In this work, we extend to \emph{dense}-frame video diffusion for improved temporal smoothness and visual fidelity.

\noindent
\textbf{Orbital Video Generation. }In contrast to single-image generation models, video diffusion models \cite{blattmann2023align, peebles2023scalable, yang2024cogvideox} are pretrained with large-scale video data, therefore preserve rich temporal priors which are beneficial for view consistency and generalization \cite{kwak2024vivid}. To exploit such priors, several works \cite{kwak2024vivid, peebles2023scalable, melas20243d, chen2024v3d, voleti2024sv3d, yang2024hi3d} propose to directly generate complete orbital videos. Specifically, Vivid-1-to-3 \cite{kwak2024vivid} prompts the video diffusion with a pretrained image-based NVS model \cite{objaverseXL} to guide the key frame generation. \cite{voleti2024sv3d, chen2024v3d, peebles2023scalable} propose to finetune a stable video diffusion (SVD) model \cite{blattmann2023align} for better model generalizability. Meanwhile, to further improve the visual fidelity of generated videos, \cite{melas20243d, wen2024ouroboros3d} adopt an iterative refinement framework by denoising results rendered from a reconstructed initial 3D model. Hi3D \cite{yang2024hi3d} follows a similar idea but instead uses estimated depth maps as a 2.5D prior for subsequent refinement. However, such coarse-to-fine approach is time-consuming and the quality of 3D guidance is directly coupled to the initial results, which complicates the training process. In contrast, we leverage the power of a 3D foundation model to incorporate robust and training-free 3D priors to effectively guide the video generation, which noticeably improves model generalizability and efficiency over iterative refinement frameworks. 

\begin{figure*}[htp!]
\centering
  {\includegraphics[width=0.95\textwidth]{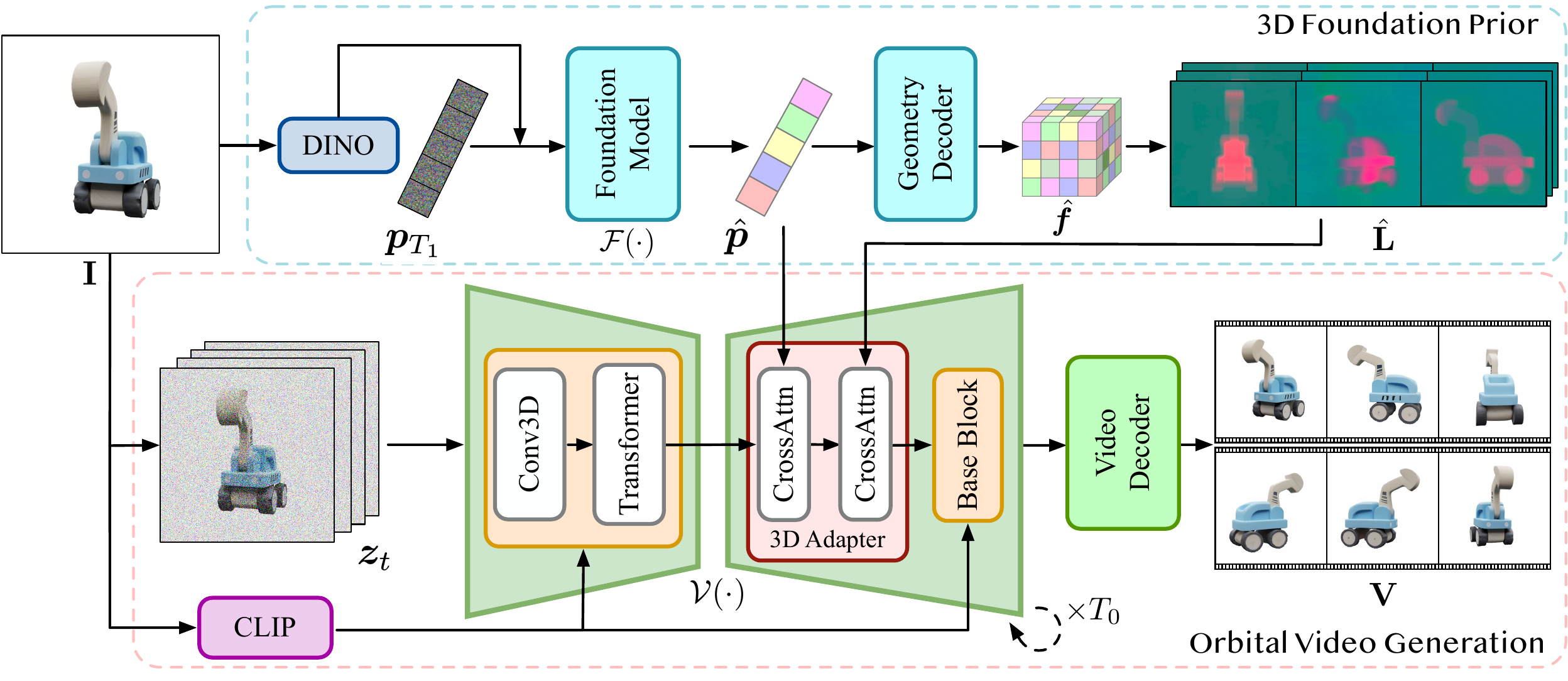}}
    \caption{\textbf{Overview of the method.} Given an input image $\mathbf{I}$, we aim to generate a realistic and consistent orbital video $\mathbf{V}$ in a base video diffusion model $\mathcal{V}(\cdot)$, conditioned on shape priors from a pretrained 3D foundation model $\mathcal{F}(\cdot)$. Specifically, we forward the input image to the foundation model to generate (\emph{i}) a denoised global latent vector $\hat{\bm{p}}$ as overall structural guidance, and (\emph{ii}) a set of latent images $\hat{\mathbf{L}}$ for view-dependent and fine-grained geometry details. We then prompt the video generation with both features through a multi-scale 3D adapter via alternating cross-attention to produce the resulting video. We omit conditions of camera pose $\bm{e}$ and timestep $t$ for simplicity.}
    \label{pipe}
\end{figure*}

\section{Methods}

\textbf{Overview. }We present our pipeline as illustrated in Figure \ref{pipe}.  Given an input object $\mathbf{I} \in \mathbb{R}^{H \times W \times 3}$, we aim to generate an orbital video $\mathbf{V} \in \mathbb{R}^{N \times H \times W \times 3}$ with $N$ frames using a base video diffusion model (Section \ref{sec31}). We assume the target camera trajectory is defined by a sequence of elevation $\{e_i\}_{i=1}^N$ and
azimuth $\{a_i\}_{i=1}^N$ angles relative to the input image, which also serves as the starting reference frame.  To encourage a realistic shape for unobserved object parts and enforce consistency across all frames, we forward the input image to a 3D foundation model to obtain its latent features as a shape prior (Section \ref{sec32}). Finally, we introduce a multi-scale 3D adapter to efficiently inject the prior into the base model (Section \ref{sec33}) for effective shape conditioning.

\subsection{Base Video Diffusion Model}
\label{sec31}
We adapt SVD \cite{blattmann2023align} as our base model for video generation, considering its simple network design and impressive performance in term of temporal consistency and generalizability. Specifically, we encode the input image as a latent embedding $\bm{z}_I \in \mathbb{R}^{H_z \times W_z \times C}$ using the VAE pretrained from SVD, and stack it with the video latent $\bm{z}_t \in \mathbb{R}^{N \times H_z \times W_z \times C}$ to construct a noisy latent with added Gaussian noise, where $H_z$ and $W_z$ represent the spatial resolution of the latent space. Next, we feed the noisy latent to a video diffusion model $\mathcal{V}(\cdot)$, constructed as a U-Net \cite{ronneberger2015u} with a sequence of residual blocks composed of Conv3D layers, followed by a series of transformer blocks with alternating spatial and temporal attention layers \cite{esser2024scaling}. In addition, we encode the CLIP \cite{radford2021learning} embedding $\bm{c}_I$ of the input image as global semantic guidance. Furthermore, we follow \cite{voleti2024sv3d} to embed all camera pose angles $\bm{e} \in \mathbb{R}^N$ and $\bm{a} \in \mathbb{R}^N$ into sinusoidal position embeddings, and then concatenate their linearly transformed results with the embedding of noise timestep $t$ to control video generation along the given camera trajectory. Both conditions are also forwarded to the video diffusion model as key-value pairs in cross-attention layers. In summary, the video diffusion model learns to denoise the noisy latent input given the above conditions as,
\begin{equation}
\hat{\bm{\epsilon}}_t = \mathcal{V}(\bm{z}_{t}| \bm{z}_I, \bm{c}_I, \bm{e}, \bm{a}) \quad t \in \{0, \cdots, T_0\}\;,
\end{equation}
where $\hat{\bm{\epsilon}}_t$ is the predicted noise. The video diffusion model is trained with the standard denoising objective, \emph{i.e.} MSE loss between the predicted and ground truth noise $\bm{\epsilon}$ as,
\begin{equation}
    \mathcal{L} = \mathbb{E}_{\mathbf{I}, \bm{\epsilon}, t}[w(t)||\mathcal{V}_{\sigma}(\bm{z}_t) - \bm{\epsilon}||_2^2]\;,
\end{equation}
where $w(t)$ is a weighting factor, and $\sigma$ is the preconditioning constants defined by \cite{karras2022elucidating}. While the above base model achieves accurate camera control, we observe that fine-tuning with only synthetic videos does not yield a robust model, and therefore often results in degraded quality when generalized to unseen objects. Moreover, using only image embeddings from a single-view input does not impose sufficient constraints on the unobserved parts of the object, hence the model often produces unrealistic structures under large viewpoint changes. To resolve both issues, we propose to augment the base model with 3D foundation priors to increase its robustness and generalizability.

\subsection{3D Foundation Priors}
\label{sec32}

To model realistic object shape distributions, we incorporate shape priors from a 3D foundation generative model trained on large-scale 3D asset corpora, enabling reconstruction from a single image with \emph{strong generalizability} (more background is provided in the appendix). In particular, we use Hunyuan3D \citep{zhao2025hunyuan3d} as the choice of foundation model, which follows a native 3D generation architecture \citep{zhang20233dshape2vecset, li2024craftsman} with shape priors well suited to our task for two key reasons: (\emph{i}) unlike previous works \cite{hong2023lrm, tang2024lgm, wu2024unique3d, wang2024crm} for 3D shape reconstruction, it does not rely on an intermediate NVS step and directly models complete object shapes in a \emph{3D latent space} \cite{chen2024dora} via tokenized point clouds, (\emph{ii}) it disentangles object shape and appearance with explicit geometry supervision, therefore results in a semantically meaningful latent space with rich object structural information. 

To exploit the power from the 3D foundation model, we propose to forward the input image to extract two scales of features as shape priors. Specifically, we first extract visual features $\bm{d}_I$ from DINOv2 \cite{oquab2023dinov2} as a patched embedding of the input image, and then condition on it to denoise a Gaussian noise vector $\bm{p}_{T_1} \in \mathbb{R}^{L \times D}$ with $L$ tokens in a rectified flow model \cite{esser2024scaling} $\mathcal{F}(\cdot)$ as,
\begin{equation}
    \hat{\bm{p}}_{t-1} = \mathcal{F}(\bm{p}_t | \bm{d}_I, t) \quad t \in \{0, \cdots, T_1\} \;.
\end{equation}
We use the final denoised vector $\hat{\bm{p}}_0$ as the latent representation of the object shape. Since the size of the shape latent is significantly smaller than the video latent, in practice the feature extraction from a separate diffusion process can be efficiently performed, as analyzed in Section \ref{section43}. While the denoised vector provides a \emph{global} structural guidance for the target object, the object geometry details are compressed as its token length is not sufficiently large. To recover \emph{local} geometry details, we first regularly partition the canonical 3D latent space into grid vertices $\bm{q} \in \mathbb{R}^{G^3 \times 3}$, where $G$ is the spatial resolution of the grid. Next, we query the grid vertices with key-value pairs encoded from the global shape vector to obtain a volumetric feature $\hat{\bm{f}} \in \mathbb{R}^{G^3 \times D'}$ from a geometry decoder $\mathcal{G}(\cdot)$ as,  
\begin{equation}
    \hat{\bm{f}} = \mathcal{G}(\text{PE}(\bm{q}); \hat{\bm{p}}_0) \;,
\end{equation}
where $\text{PE}(\cdot)$ is the positional embedding for the grid vertices, and we follow \cite{zhang20233dshape2vecset} to use a cross-attention layer as the geometry decoder. As the volumetric features are interpreted as SDF values in the foundation model, they are explicitly supervised to encode local geometry details for fine-grained object representation. In summary, the global and local features combined summarize the complete object shapes and serve as an effective shape condition.

\subsection{Multi-Scale 3D Adapter}
\label{sec33}
Motivated by \cite{ye2023ip}, we introduce a novel multi-scale 3D adapter as an effective plug-in module to inject 3D foundation priors into the base video model. Specifically, for the global latent vector, we first duplicate it into $N$ batches $\hat{\bm{p}} \in \mathbb{R}^{N \times L \times D}$ to share all frames a unified shape reference. Next, for each input feature $\mathbf{f}_i^{(0)} \in \mathbb{R}^{N \times (H_z \times W_z) \times C}$ of the $i$-th transformer block, we fuse it as the query with key-value pairs embedded by the global vector in a cross-attention layer as,
\begin{equation}
    \mathbf{f}_i^{(1)} = \mathbf{f}_i^{(0)} + \text{CrossAttn}(\mathbf{f}_i^{(0)}; \text{MLP}(\hat{\bm{p}})) \;,
\end{equation}
where $\text{MLP}(\cdot)$ aligns the channels between the two features. As studied in \cite{lai2025unleashing}, vectors denoised from vecset diffusion are locality-preserving similar to patch features from vision transformers \cite{caron2021emerging}, thus individual tokens represent meaningful geometry semantics compatible with the attention mechanism. 

To provide the base video model with view-dependent guidance, we project the volumetric feature $\hat{\bm{f}}$ into $M$ canonical views with predefined camera extrinsics (see Section \ref{section42} for more details). Instead of using all $N$ views, we find these canonical views are sufficiently representative to describe the object shape, while significantly reducing the memory overhead. Moreover, since the object inferred by the 3D foundation model may not perfectly align in the orientation with the target object, we propose to implicitly learn such discrepancy via 3D attention layers \cite{shi2023mvdream}. Specifically, we first forward the latent images to a Conv2D encoder $f_e(\cdot)$ to obtain downsampled results $\hat{\mathbf{L}} \in \mathbb{R}^{M \times H_l \times W_l \times D'}$. We then fold all views into tokens (followed by expanded batches) as $\hat{\bm{l}} \in \mathbb{R}^{N \times (M \times H_l \times W_l) \times D'}$, which can be used as a key-value condition in a separate cross-attention layer as,
\begin{equation}
    \mathbf{f}_i^{(2)} = \mathbf{f}_i^{(1)} + \text{CrossAttn}(\mathbf{f}_i^{(1)}; \text{MLP}(\hat{\bm{l}})) \;,
\end{equation}
where the final output $\mathbf{f}_i^{(2)}$ will be processed by the following blocks in the base video model. In this way, we augment the base video model with multi-scale shape features, which are critical to encourage realistic and coherent results with improved model robustness and generalizability. Moreover, by using latent features as object shape representations, we avoid the time-consuming mesh extraction process \cite{lorensen1998marching} and ensure a more efficient inference pipeline.

\section{Experiments}

\begin{table*}[t]
\centering
\caption{\textbf{Quantitative comparison with baseline methods.} Best results are marked as \textbf{bold}. Our method achieves superior results than baseline methods \cite{long2024wonder3d, li2024era3d, voleti2024sv3d, yang2024hi3d, zhao2025hunyuan3d, xiang2024structured} in all visual quality, shape realism and view consistency metrics on multiple benchmarks \cite{objaverseXL, downs2022google}. }
\label{tab:garment_deformation}
\resizebox{0.95\textwidth}{!}{
\begin{tabular}{@{}lcccccccccc@{}}
\toprule
\multirow{3}{*}{Methods} & \multicolumn{5}{c}{Objaverse-XL \cite{objaverseXL}} & \multicolumn{5}{c}{GSO \cite{downs2022google}} \\
\cmidrule(l){2-6} \cmidrule(l){7-11} & PSNR $\uparrow$ & SSIM $\uparrow$ & LPIPS $\downarrow$ &
                 CLIP-S $\uparrow$ &
                 MEt3R $\downarrow$ &
                  PSNR $\uparrow$ & SSIM $\uparrow$ & LPIPS $\downarrow$ &
                 CLIP-S $\uparrow$ &
                 MEt3R $\downarrow$  \\ \midrule
Wonder3D \citep{long2024wonder3d} & 19.53 & 0.89 & 0.15 & 89.03 & - &  19.06 & 0.89 & 0.19 & 84.59 & -\\
Era3D \cite{li2024era3d} & 21.08 & 0.90 & 0.13 & 91.06 & - & 21.72 & 0.91 & 0.16 & 88.11 & -\\
Ours (6 frames) & \textbf{23.20} & \textbf{0.93} & \textbf{0.08} & \textbf{94.95} & - & \textbf{25.05} & \textbf{0.93} & \textbf{0.07} & \textbf{92.86} & -\\ \midrule
Trellis \citep{xiang2024structured} & 19.43 & 0.89 & 0.14 & 92.05 & - & 20.36 & 0.91 & 0.15 & 89.70 & -  \\ 
Hunyuan \citep{zhao2025hunyuan3d} &  20.25 & 0.91 & 0.11 & 93.44 & - & 20.88 & 0.91 & 0.14 & 91.03 & -  \\
SV3D \citep{voleti2024sv3d} & 20.48 & 0.91 & 0.12 & 92.84 & 0.07 & 21.58 & 0.92 & 0.12 & 89.45 & 0.06\\ 
Hi3D \citep{yang2024hi3d} & 19.32 & 0.90  & 0.14 & 90.61 & 0.09 & 19.76 & 0.90 & 0.16 & 87.56 & 0.07 \\
Ours (21 frames) & \textbf{22.78} & \textbf{0.92} & \textbf{0.09} & \textbf{94.19} & \textbf{0.05} & \textbf{24.67} & \textbf{0.93} & \textbf{0.08} & \textbf{91.90} & \textbf{0.04}  \\
\bottomrule
\end{tabular}
}
\label{table1}
\end{table*}

\subsection{Datasets} To improve model generalizability, we train our model on a large-scale dataset Objaverse-XL \cite{objaverseXL} with high-quality artist-created 3D assets, and follow the same pipeline from \cite{voleti2024sv3d} to render synthetic orbital videos as ground truths. More details about the training data are included in the appendix. For evaluation, we use 150 unseen objects from Objaverse-XL and 100 unseen objects from the GSO \cite{downs2022google} dataset filtered to ensure no duplication.


\subsection{Metrics}
We measure and compare the quality of generated videos in three aspects: visual fidelity, shape realism, and geometry consistency. For visual fidelity, we follow standard metrics in \cite{voleti2024sv3d} to evaluate Peak Signal-to-Noise Ratio (PSNR), Structural Similarity (SSIM), and Learned Perceptual Similarity
(LPIPS) \cite{zhang2018unreasonable}, between generated and ground truth frames. For shape realism, we compute the CLIP-score (CLIP-S) \cite{radford2021learning} to evaluate the semantic alignment against image distribution with realistic shapes. 

Unlike 3D generation methods \citep{zhao2025hunyuan3d, xiang2024structured} that aim to recover complete 3D shapes, our work primarily focuses on generating realistic orbital videos for their \emph{visual aesthetics}. Moreover, subsequent reconstruction from the generated RGB videos is non-trivial, and the final geometry quality heavily depends on the chosen reconstruction method and desired resolution. To this end, evaluation with 3D metrics such as Chamfer distances is unfair and does not directly reflect the geometric quality of the generated videos. Alternatively, we adopt the recent metric MEt3R \citep{asim2025met3r} by computing  it between all neighboring frame pairs in the generated videos to directly evaluate consistency in the pixel space.

\subsection{Implementation Details}
\label{section42}
We implement our model in PyTorch \cite{imambi2021pytorch} and perform the experiments on 8 NVIDIA H200 GPUs. We train the model using the
Adam \cite{kingma2014adam} optimizer for 80K iterations with a learning rate of $1\times10^{-5}$ at an effective batch size of 16. For the base video model, we follow \cite{voleti2024sv3d} to generate 21-frame videos with 576$\times$576 pixels. For the 3D foundation model, we use 3072 tokens for the global vector as well as a grid resolution of $128^3$ for the volumetric feature. The rectified flow model and geometric decoder are both initialized from pretrained weights in \cite{zhao2025hunyuan3d} considering its promising generalizability, and remain frozen during training and inference. We project latent images with $M = 8$ canonical views, \emph{i.e.} left, right, front, rear and their diagonal views, for static orbits (0 elevation). For dynamic orbits where the elevation angles vary between [-60, 60], we further include top and bottom views to ensure sufficient view coverage. The final resolution of the latent images is $16 \times 16$ after downsampled by $f_e(\cdot)$. During inference, we perform $T_0 = 50$ diffusion steps for the base video model and $T_1 = 30$ steps for obtaining the multi-scale features. Finally, we follow the CFG \cite{ho2022classifier} strategy in  \cite{voleti2024sv3d} for inference, and nullify all features from the foundation model for unconditioned generation.

\subsection{Results and Comparisons}
\label{section43}
\textbf{Quantitative Comparison.} We compare with three types of state-of-the-art baselines: (\emph{i}) orbital video generation methods SV3D \cite{voleti2024sv3d} and Hi3D \cite{yang2024hi3d} which are directly comparable to our setting, (\emph{ii}) NVS methods Wonder3D \cite{long2024wonder3d} and Era3D \cite{li2024era3d} that generate images on a predefined set of sparse canonical views, and (\emph{iii}) 3D generation methods Hunyuan3D-2 \cite{zhao2025hunyuan3d} and Trellis \cite{xiang2024structured} by rendering orbital videos from reconstructed textured meshes. Since \cite{long2024wonder3d, li2024era3d, yang2024hi3d} output different numbers of frames, we manually select frame indices which best align with the ground truths views for a fair comparison. For \cite{xiang2024structured, zhao2025hunyuan3d}, we align the orientation of generated meshes with the ground truth meshes and render their results using ground truth camera poses. Since both methods bake lighting from input images into their texture, we directly rasterize their results to fairly evaluate their texture quality. Note that we only compare MEt3R metrics for video generation works \cite{voleti2024sv3d, yang2024hi3d} as the neighboring frames contain sufficient overlapping. As shown in Table \ref{table1}, our method outperforms baselines by a large margin in both visual quality and semantic alignment, leading to more realistic results. Moreover, compared to existing video generation methods \cite{voleti2024sv3d, yang2024hi3d}, our method also demonstrates superior multi-view consistency thanks to the utilization of a unified 3D reference. 

\begin{figure*}[t]
    \centering
    \includegraphics[width=1.0\linewidth]{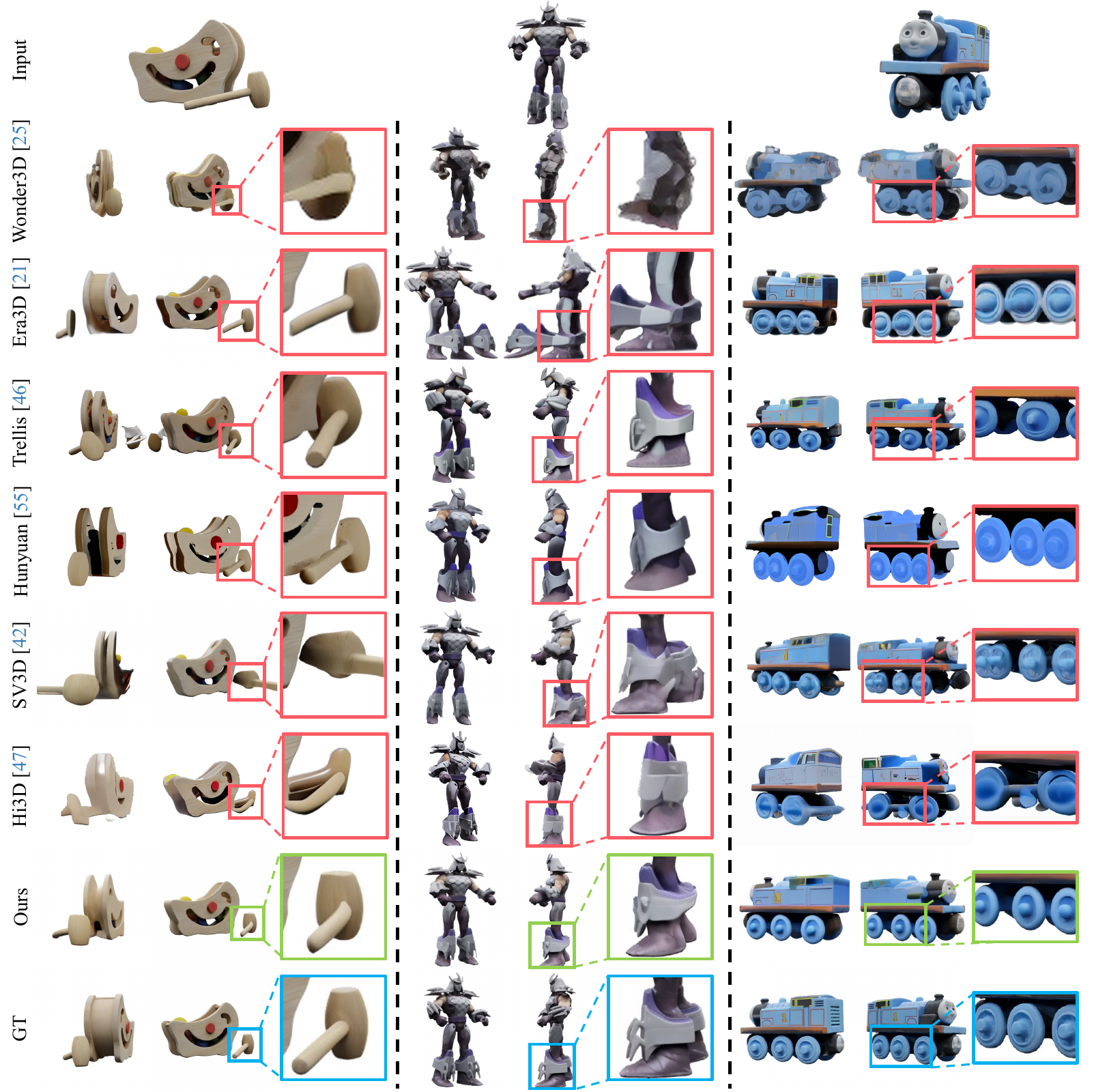}
    \vspace{-0.4cm}
    \caption{\textbf{Qualitative comparison with baselines.} Compared to NVS \cite{long2024wonder3d, li2024era3d} and video \cite{voleti2024sv3d, yang2024hi3d} generation works, our method produces more realistic and consistent results with less artifacts, \emph{e.g.} distorted shapes and unnatural structures. Moreover, we generate more accurate object colors and lighting effects compared to rendered textured meshes from 3D generation methods \cite{xiang2024structured, zhao2025hunyuan3d}.}
    \label{quali}
    \vspace{-0.4cm}
\end{figure*}

\noindent
\textbf{Qualitative Comparison.} We further compare qualitative results with baselines in Figure \ref{quali}. For image and video generation methods \cite{long2024wonder3d, li2024era3d, voleti2024sv3d, yang2024hi3d}, due to the lack of explicit prior knowledge from 3D geometry supervision, they often produce results with undesired artifacts, such as distorted shapes and unnatural structures, especially when synthesizing challenging side and rear views. In contrast, we achieve superior shape realism and consistency thanks to the guidance of shape priors from the foundation model. Moreover, while 3D generation methods \cite{xiang2024structured, zhao2025hunyuan3d} can directly reconstruct coherent 3D shapes, we observe they often fail to recover accurate textures that faithfully preserve object appearance and view-dependent lighting effects. As a result, they can produce visually degraded results such as  misaligned base colors, \emph{e.g.} last example for \cite{zhao2025hunyuan3d}, or blurred shadows. In addition, we observe \cite{zhao2025hunyuan3d} can produce object meshes with higher quality over the alternative foundation model \cite{xiang2024structured} without producing distorted shapes, which motivates us to adopt it for effective shape priors. Finally, we refer readers to more qualitative comparisons with other methods in the appendix.

\noindent
\textbf{Generalization Evaluation.} To evaluate the generalizability of our model, we show in Figure \ref{gen} that our method can be applied to in-the-wild images with various types of objects and input views, while consistently maintaining realistic shapes and noticeably achieving better visual quality than previous video generation methods \cite{voleti2024sv3d, yang2024hi3d}. While we benchmark all results on static orbits with zero elevations, we show in Figure \ref{dynamic} that our method can be extended to generate complex dynamic orbits \cite{voleti2024sv3d}, which demonstrates its capability for accurate camera control.

\noindent
\textbf{Inference Time Analysis. }Thanks to our optimized implementation by \cite{zhao2025hunyuan3d, lai2025unleashing}, the overall time to denoise the global latent vector takes only 1.8 sec per input image. The query and projection time for generating local volumetric features and latent images further take 0.34 sec and 0.11 sec respectively. Finally, the multi-scale 3D adapter requires additional 0.28 sec per sampling step during inference, and contains in total 0.3B trainable parameters. All experiments are performed on a single NVIDIA H200 GPU. In summary, the proposed modules do not introduce significant overhead.

\begin{figure*}[t]
    \centering
    \includegraphics[width=0.95\textwidth]{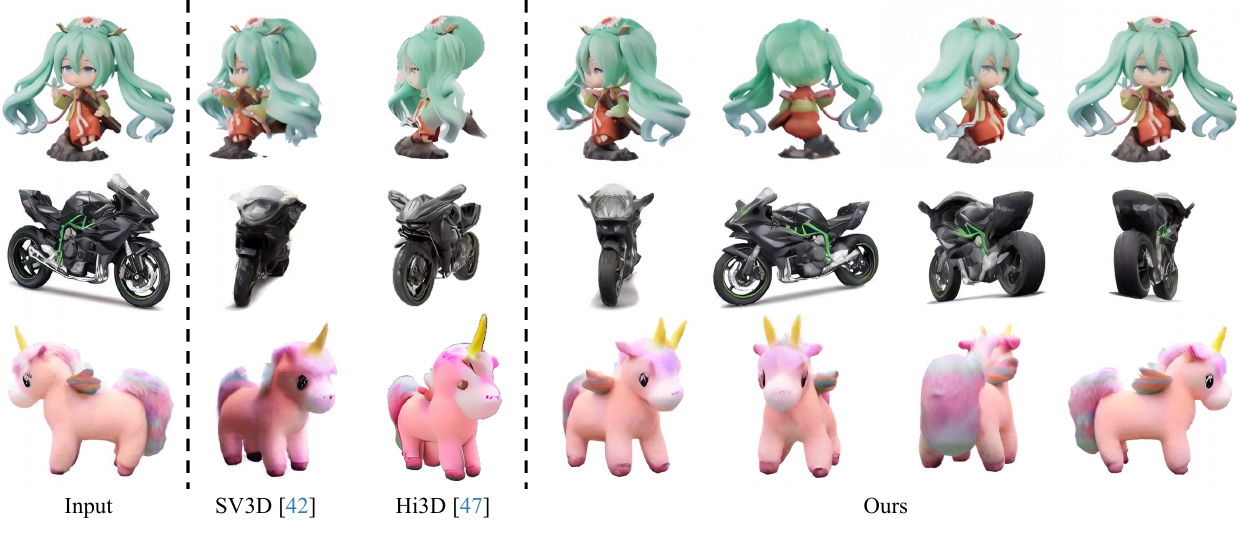}
    \vspace{-0.3cm}
    \caption{\textbf{Results on in-the-wild images.} Our method robustly generalizes to in-the-wild examples and outperforms baseline video generation methods \cite{voleti2024sv3d, yang2024hi3d} in both visual fidelity and shape realism.}
\end{figure*}

\begin{figure*}[t]
    \centering
    \includegraphics[width=0.95\textwidth]{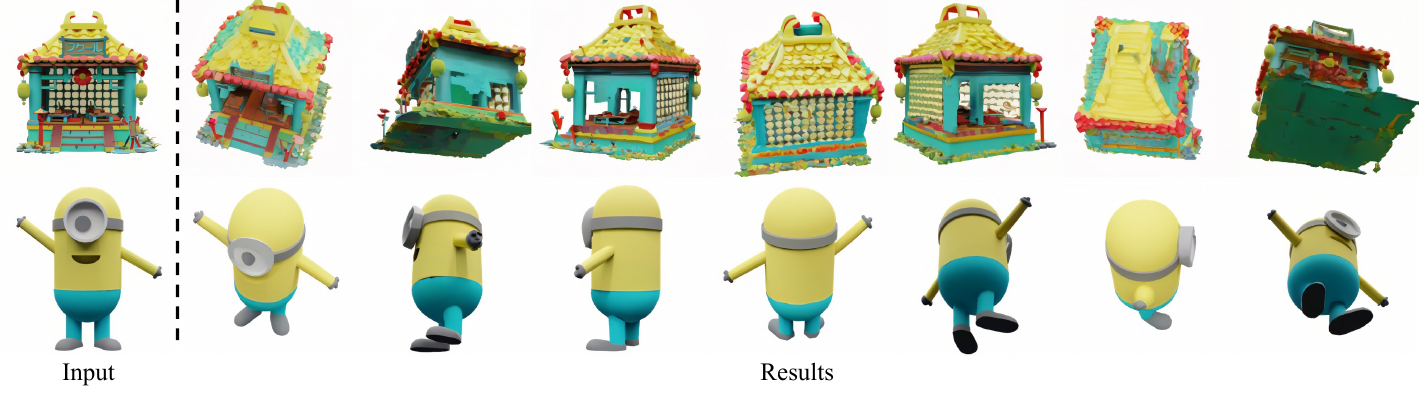}
    \vspace{-0.3cm}
    \caption{\textbf{Extension to dynamic orbits.} Our method can follow complex camera trajectories with non-zero elevations, \emph{i.e.} dynamic orbits defined by \cite{voleti2024sv3d} to achieve accurate camera controls.}
    \label{dynamic}
    \vspace{-0.4cm}
    \label{gen}
\end{figure*}

\subsection{Ablation Study}

\textbf{Effects of 3D Foundation Priors.} In Table \ref{table2}, we show the effects of the proposed 3D foundation priors. For a fair comparison, we first retrain the SV3D model using our dataset partition and obtain a comparable base model (w/o Priors). We observe that incorporating global latent vector noticeably improves the multi-view consistency (lower MEt3R metric) and shape realism (higher CLIP-S metric). Moreover, as the volumetric features contain rich view-dependent and fine-grained local geometry details, we find that including both scales of features (Global + Local) in our model further improves the overall performance. 

\begin{figure}[t]
    \centering
    \includegraphics[width=0.5\textwidth]{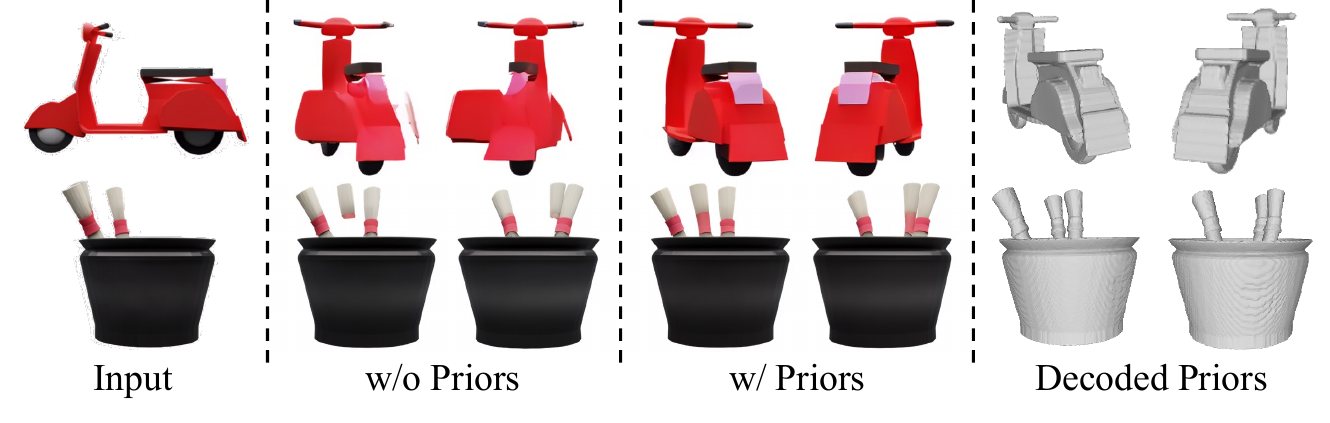}
    \vspace{-0.7cm}
    \caption{\textbf{Effects of the 3D foundation priors.} The shape priors from 3D foundation models (visualized after decoding and mesh extraction at the last column) help to provide complementary guidance for unobserved and occluded object parts.}
    \label{ablation}
    \vspace{-0.3cm}
\end{figure}

In addition, we show in Figure \ref{ablation} for qualitative analysis of the 3D foundation priors. For base model without priors, we often observe blurry and unrealistic structures for object parts unobserved from input image. Moreover, for complex scenes with mutual occlusion between objects, it is challenging for the base model alone to analyze spatial relationships, consequently producing results with incorrect shapes. In comparison, by leveraging 3D foundation priors as complementary guidance (visualized after explicit mesh extraction from the predicted volumetric features using Marching Cubes \cite{lorensen1998marching}), our method achieves results with superior quality and authentic shapes. Meanwhile, by leveraging the attention-based adapter as a \emph{soft} constraint, the base video model preserves its stochastic nature and capability to balance between image and shape conditions. We refer readers to more ablation studies for this in the appendix.

\begin{table}[t]
\centering
\caption{\textbf{Ablation studies.} In top rows, we observe leveraging both scales of
features achieves the best
overall performance. In bottom rows, we show proposed cross-attention based adapter outperforms alternative feature conditioning approaches as it better preserves pretrained temporal priors.}
\resizebox{0.48\textwidth}{!}{
\begin{tabular}{@{}lccccc@{}}
            \toprule
            Methods & PSNR $\uparrow$ & SSIM $\uparrow$ & LPIPS $\downarrow$ & CLIP-S $\uparrow$ & MEt3R $\downarrow$ \\ \midrule
            w/o Priors & 20.06 & 0.90 & 0.14 & 91.26 & 0.08 \\
            w/ Global & 21.86 & 0.91 & 0.11 & 93.51 & 0.06 \\
            Global + Local & \textbf{22.78} & \textbf{0.92} & \textbf{0.09} & \textbf{94.19} & \textbf{0.05} \\ \midrule

            Feature Cat. &  20.53 & 0.90  & 0.16  & 91.88 & 0.08 \\
            Frame Stack & 19.27 & 0.89 & 0.18 & 89.96 & 0.08 \\
            CrossAttn & \textbf{22.78} & \textbf{0.92} & \textbf{0.09} & \textbf{94.19} & \textbf{0.05} \\ \bottomrule
        \end{tabular}}
\label{table2}
\vspace{-0.5cm}
\end{table}

\noindent
\textbf{Design Variants of 3D Adapter.} We compare with other two variants of designs for the 3D adapter: (\emph{i}) we perform average pooling on the tokens of the global vector to produce a 1D feature $\hat{\bm{p}}' \in \mathbb{R}^{D}$, then concatenate it with other features such as camera poses as condition (Feature Cat.) (\emph{ii}) We render all $N$ latent images and stack it with the noisy latents $\bm{z}_t$ channel-wise as new inputs to the video diffusion model (Frame Stack). As shown in Table \ref{table2}, we find both designs lead to inferior performance compared to the cross attention approach (CrossAttn). In particular, projecting the global latent into a 1D vector destructs its encoded geometry information, which is not beneficial for effective shape conditioning. In addition, since the latent spaces of 3D and image features differ a lot, stacking them together as input will significantly alter the distribution of the pretrained video diffusion model, which negatively impacts the fine-tuning process. In comparison, the cross attention approach best preserves the capability of the base video model. We find this design also leads to a simpler training process and thus helps to improve overall model performance.

\section{Discussion}
\label{section5}
\textbf{Limitation.} Although our method achieves results with superior consistency and shape realism, the visual fidelity of the generated videos is
bounded by the resolution of the base video model, thus may lead to degraded quality for tiny details. In addition, we only leverage shape priors from the foundation model as we observed its texture quality is not sufficiently robust, hence synthesizing complex unobserved object appearances leveraging 3D foundation priors remains unexplored. Finally, as variants of 3D foundation models \cite{lai2025unleashing, xiang2024structured} can enable faster diffusion than video models, we encourage future works to further improve efficiency.

\noindent
\textbf{Conclusion.} In this paper, we presented a novel pipeline for generating orbital videos with more realistic object shapes and consistent frames by incorporating 3D foundation priors. To guide the synthesis of unseen object parts under large viewpoint changes, we leverage shape priors learned by a 3D foundation model as an auxiliary constraint, which noticeably improves shape realism in the results. Furthermore, we introduce a multi-scale 3D adapter that effectively conditions the video generation with global and local latent features. These contributions lead to improved visual fidelity and robustness, while also enabling potential extensions to more general video generation tasks.

{
    \small
    \bibliographystyle{ieeenat_fullname}
    \bibliography{main}
}

\newpage


\section*{Supplementary material}

## A. Background for 3D Foundation Models

Foundation models for learning object geometry priors have been widely studied via image representations, such as depth [2, 19], normal [1, 9, 15] or point [26, 27] maps. While these models mostly describe object structures for input view only, recent *3D foundation models* learn to directly encode object geometry in a native 3D latent space, *e.g.* constructed by tokenized point clouds [4, 34] or voxels [17, 20], thus can model complete object shapes. Specifically, [16, 35, 36] adopt a 3DShape2VecSet [34] architecture to compress sampled object points as a latent vector, which can be efficiently denoised via rectified flow models [8, 18]. Alternatively, [17, 30] propose a two-stage diffusion process using sparse voxels to support detailed and localized shape generation. Moreover, to avoid time-consuming mesh extraction and support real-time rendering, several works [22, 28, 31] adopt Gaussian Splatting [14] as the 3D representation and learns to decode corresponding parameters during inference, which enables the model to capture more realistic lighting effects in the rendered results. Unlike previous pipelines for 3D shape generation [10, 11, 21, 24, 29] that rely on an intermediate multi-view synthesis step, above methods fully operate on 3D object representations with explicit 3D geometry supervision, thus effectively model realistic object shape distributions as foundation priors.

## B. Data Processing Pipeline

We use a subset of Objaverse-XL [7] with 160K assets as our training data and follow [25] to render synthetic orbital videos as ground truths. Specifically, we first normalize each object inside a unit cube and center it at the origin. Note that such normalization is also assumed by the 3D foundation model. We then use Blender's CYCLES engine to render 84-frame orbits of 576×576 pixels images with white background. During training, we randomly select a frame as the starting frame with a step of 4 to construct the target 21-frame orbit. For lighting, we follow the script from [30] to illuminate the object with curated HDRI environment maps. Finally, we follow [25] to set the camera FoV as 33.8 degrees. For static orbits, we sample azimuth angles that equally divide 360 degrees, and set the elevation to 0. For dynamic orbits, we smoothly transition the elevation values in the range [-60, 60] using a random set of sinusoids. In summary, our data processing pipeline closely follows [25], so that the finetuned base model from SVD shares a comparable performance.

## C. More Ablations for 3D Foundation Priors

In Figure 1, we show generated samples using the same input image and shape prior but different random seeds. We observe that incorporating 3D adapter preserves the stochastic nature of the base video generative model, thus enabling the generation of videos with different object appearances. Moreover, all generated results close resemble the shape prior, *i.e.* as visualized in the rightmost column, which demonstrates its capability of shape control. In addition, we observe that the shape priors only act as a *soft* constraint, where generated images can differ in fine-grained object parts. We attribute this as two reasons: (*i*) both images and 3D shapes are processed in *patches* compressed by their respective VAEs, making conditioning on individual pixels challenging, (*ii*) the image and shape priors are conditioned in parallel via cross attention layers, hence the base video model has its own capability of balancing from multiple condition signals. Similar to other adapters [33], we observe that achieving strict identity consistency is challenging for 3D shape priors as well. Nevertheless, by incorporating shape conditions as an auxiliary constraint, our method can effectively regularize the overall structure and robustly ensure a plausible output.

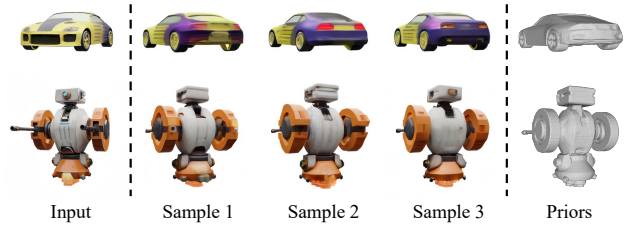

Figure 1. **Effects of random seeds for shape priors**. Our method can generate different object appearances following the same prior, preserving the stochastic nature of generative model.

In Figure 2, we visualize decoded shape priors under different seeds for the 3D foundation model. Since the foundation model uses a large CFG scale by default [36], we observe it produces a consistent shape prior with minor variance, which helps to accelerate the convergence for training and ensures a consistent behavior during inference.

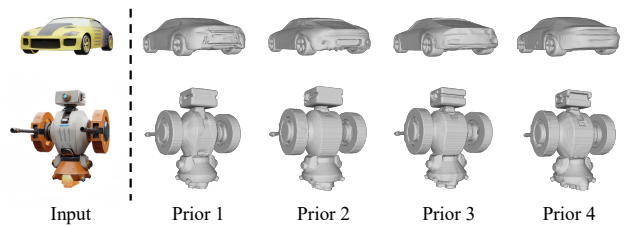

Figure 2. **Effects of random seeds for the 3D foundation model**. By incorporating a large CFG scale, the 3D foundation model demonstrate a consistent behavior on resulting priors

## D. More Results

We show in Figure 6 for results on more unseen examples with more views. Our method consistently generalizes to

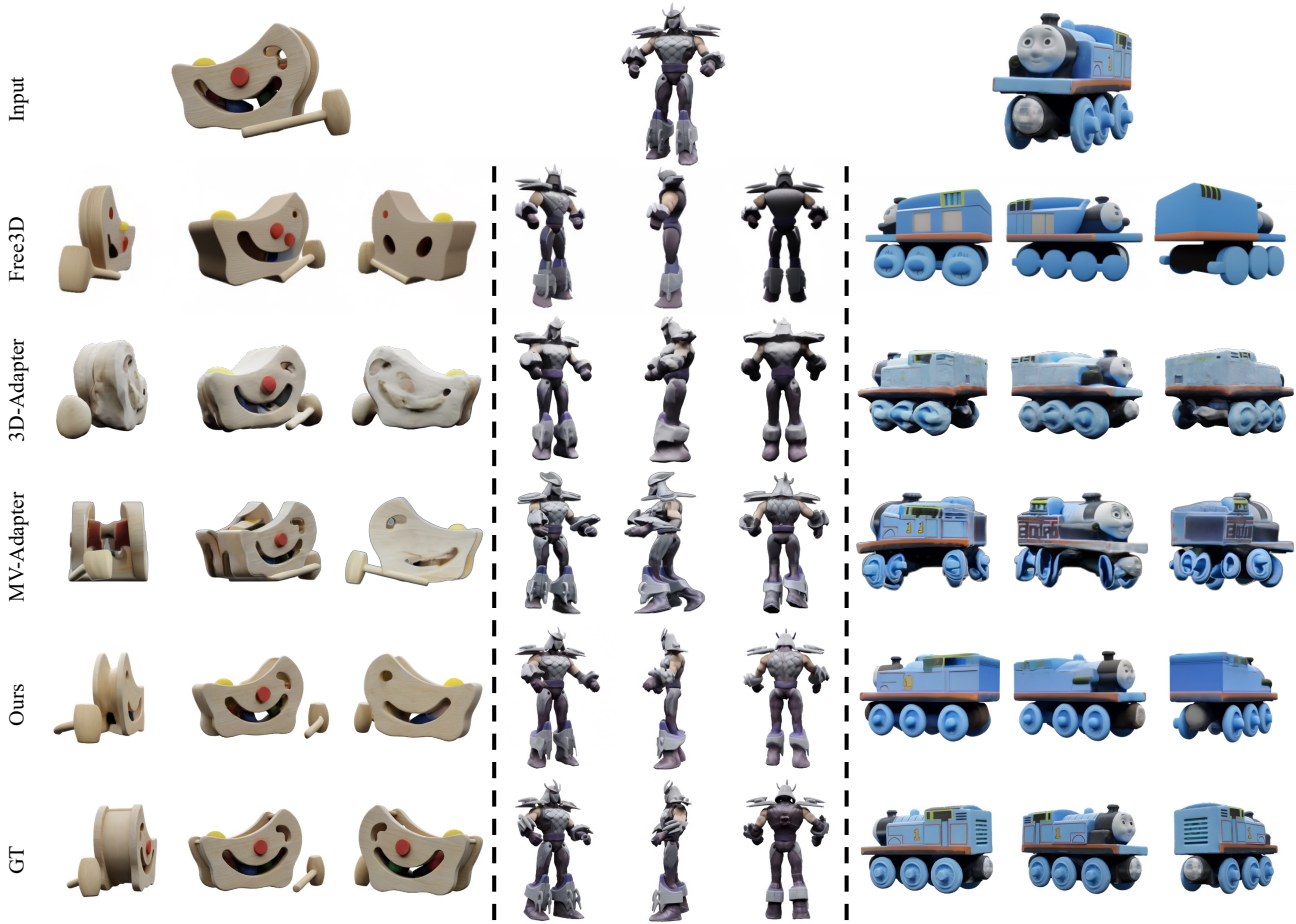

Figure 3. **Qualitative Comparisons with NVS methods [5, 13, 37] for completeness.** Our method produces more realistic results with improved fidelity in shape and appearances details. Moreover, we directly output smooth video results leveraging temporal priors from general video model [3], as shown in supplementary videos.

diverse types of objects and produces consistent frames with realistic object shapes.

## E. More Comparisons with NVS Methods

We show more comparisons with several recent NVS baselines: Free3D [37], 3D-Adapter [5, 6], and MV-Adapter [13] in Figure 3. We obtain all results using their official code and weights. Since these methods either generate individual or sparse frames during inference, they are not directly comparable with our method or baselines methods [25, 32] that target on producing temporal consistent orbital videos with dense frames. Nevertheless, qualitative comparison show that our method achieves superior fidelity in terms of both shape and appearances details, leading to results with noticeably improved realism.

## F. More Comparison with Foundation Models

We show more comparisons with several recent 3D foundation models: Ouroboros3D [28], Gen-3Diffusion (GEN3D) [31], and an improved version of Hunyuan3D (Hunyuan2.1) [23] in Figure 7. Both [28] and [31] learn to generate 3DGS as the 3D representation and renders results with view-dependent lighting effects via Gaussian Splatting. In addition, unlike Hunyuan2.0 [36] that bakes the input lighting into the textures, Hunyuan2.1 [23] first delight the input image and estimate PBR textures from the intermediate inputs, thus allows relighting for the generated meshes to produce more complex light effects such as reflection for mental materials. In the comparison, we show that our method remain superior in visual quality, especially in terms of the texture fidelity in rendered results as well as base color alignment to the ground truth. To this end, this further verifies the value of orbital video generation methods for visual aesthetic.

## G. 3D Reconstruction from Video Results

We show in Figure 4 that similar to baseline works [25, 32], our method can be extended to generate 3D meshes from the video results, which also demonstrates the multi-view con-

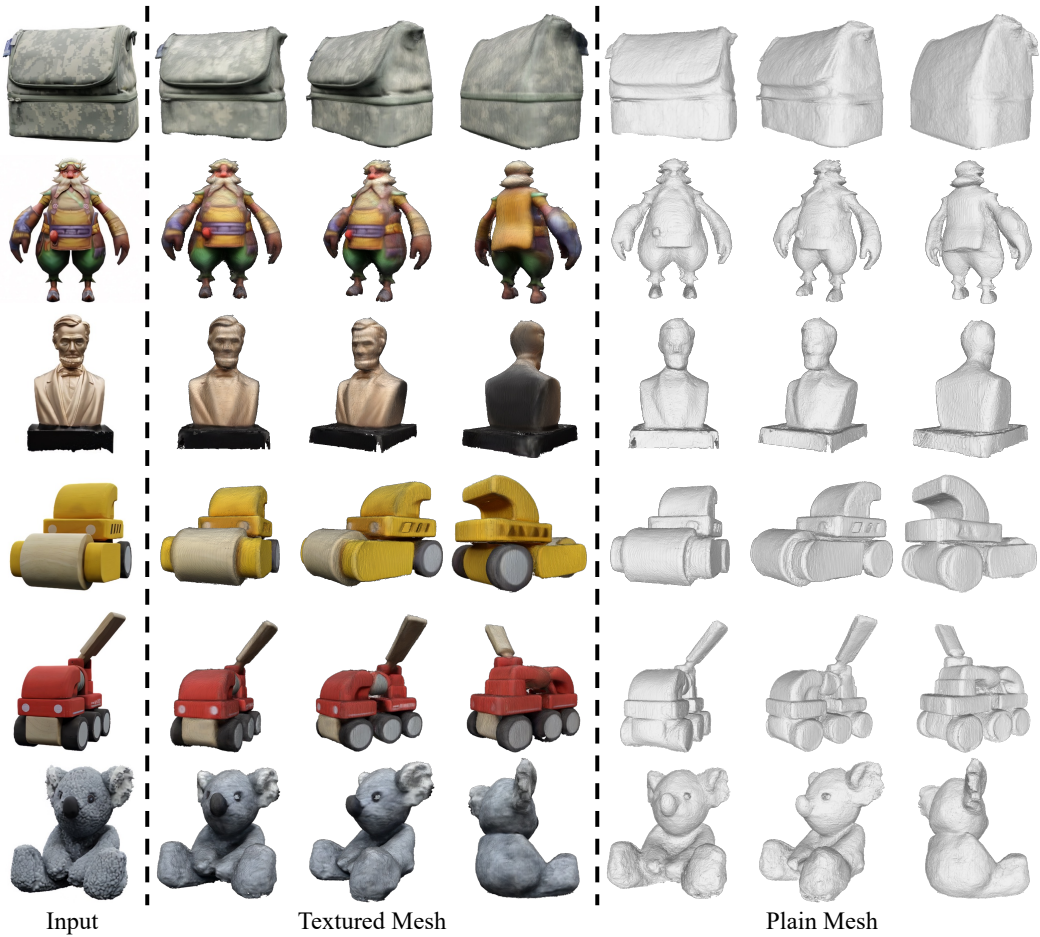

Input          Textured Mesh          Plain Mesh

Figure 4. **Reconstructed textured meshes from generated video results.** Similar to [25, 32], our method can be extended to reconstruct textured 3D meshes, which also demonstrates the multi-view consistency of generated videos.

sistency. Specifically, we use a simple method that adopts 2DGS [12] to directly reconstruct textured meshes using the predicted RGB images, and observe that the obtained 3D models faithfully recover the input images.

## H. Failure Cases

We show examples of failure cases in Figure 5. Due to the limited geometry resolution, the 3D foundation model fails to generate accurate shape priors for complex scenes and tiny object regions. Moreover, the visual fidelity of the results is also bounded by the capacity of the base video model, *e.g.* SVD used in our method, which makes synthesizing and controlling for detailed object textures and components challenging. Nevertheless, we compare with [32] with higher spatial resolution and find simply scaling the base video model without incorporating 3D shape priors does not yield improved results. In view of this, future research works are encouraged to combine the shape priors with more powerful base video models to enhance visual fidelity.

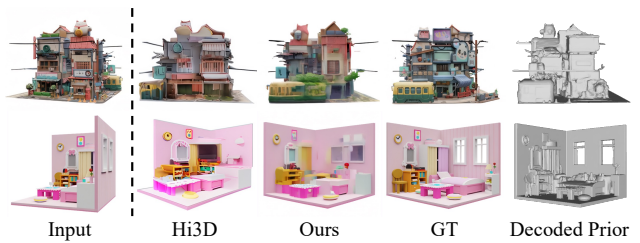

Input     Hi3D     Ours     GT     Decoded Prior

Figure 5. **Examples of failure cases.** Due to the limited resolution, both base video model and 3D foundation model fail to recover fine-grained details on complex scenes, leading to blurry results and ineffective shape control particular when the shape priors disagree with ground truth videos.

## I. Supplementary Video

We refer readers with more results in the supplementary videos, tested on a wide range of objects and real world examples to demonstrate the efficacy and generalizability of the proposed method.

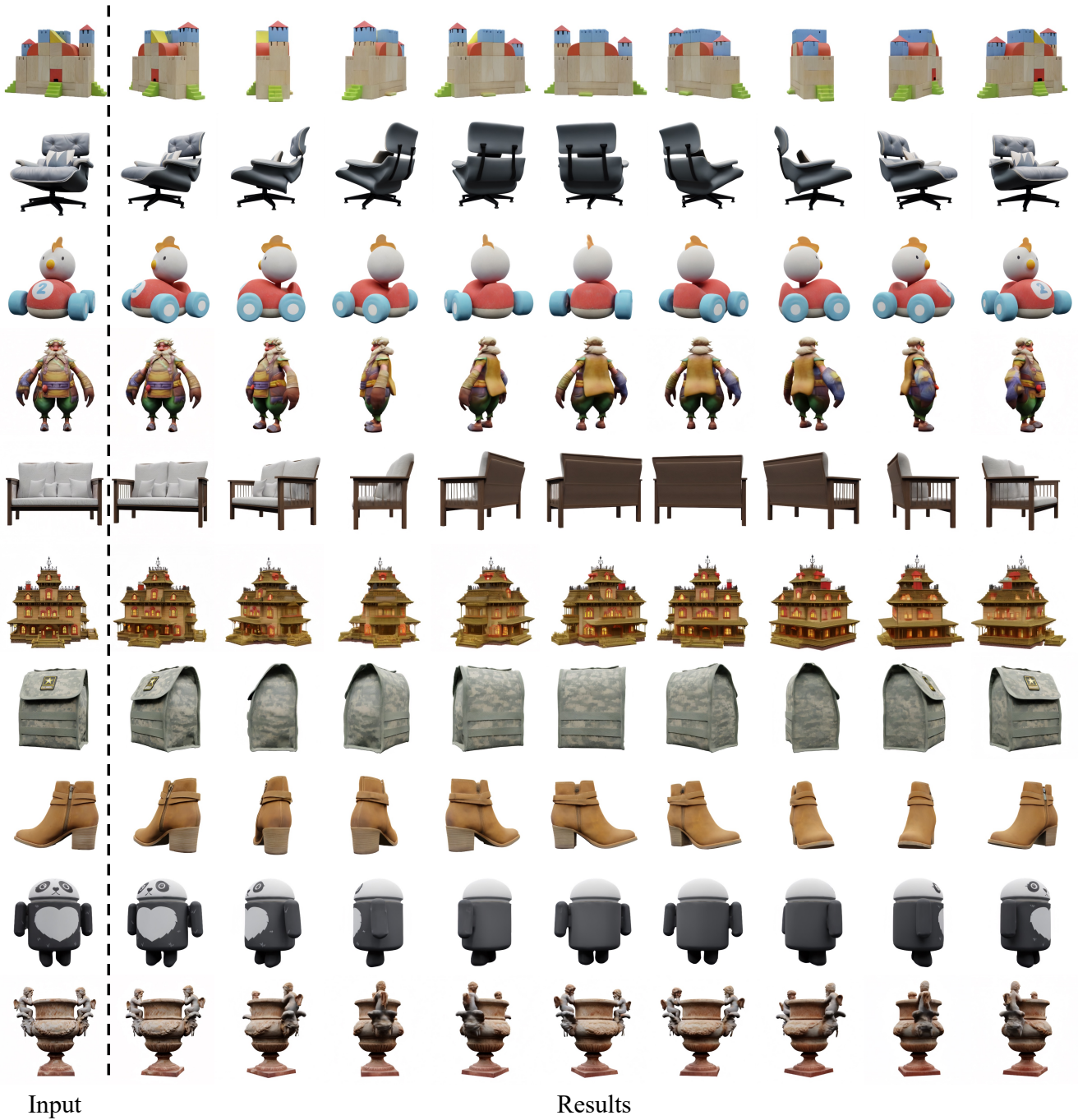

Input                    Results

Figure 6. **We show more results of generated frames on diverse input views and unseen objects.**

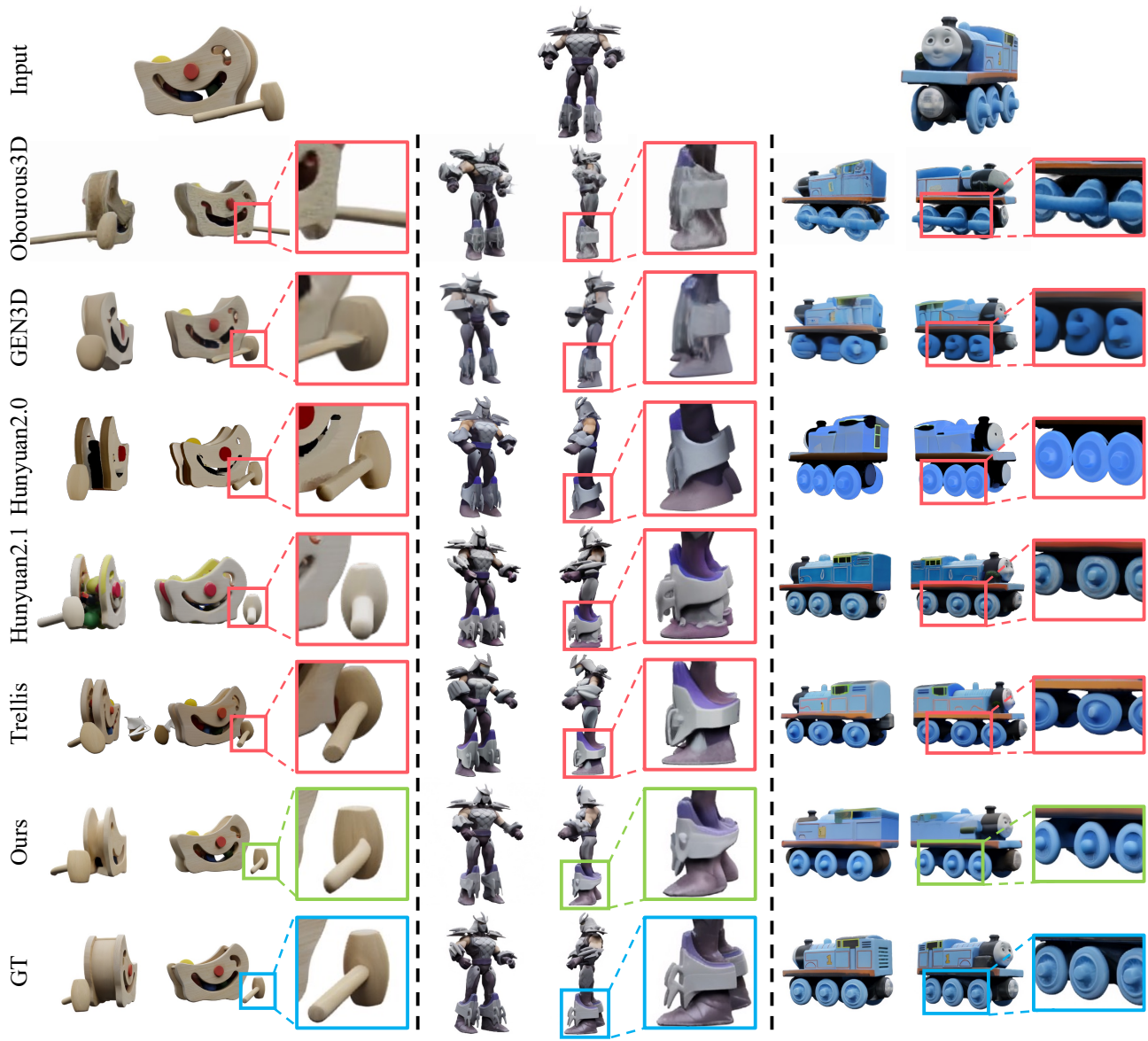

Figure 7. **Qualitative Comparisons with 3D generation methods [23, 28, 31].** Our method produces results with higher fidelity compared to 3DGS based methods [28, 31] and ensures better aligned base colors over foundation models that explicitly estimate mesh texture maps [23, 36]. This further verifies the efficacy of native video generation methods in producing visually improved results.



\end{document}